%% file: article.tex
\documentclass[times,twocolumn,final]{elsarticle}
\makeatletter\if@twocolumn\PassOptionsToPackage{switch}{lineno}\else\fi\makeatother

\title{End-to-End Semantic Video Transformer for Zero-Shot Action Recognition}

\author[1]{Keval Doshi\fnref{fn1}} 
\ead{kevaldoshi@usf.edu}
\fntext[fn1]{Performed the work while at University of South Florida.}
\author[2]{Yasin Yilmaz\corref{cor}}
\ead{yasiny@usf.edu}
\cortext[cor]{Corresponding author}

\affiliation[1]{organization={Amazon}, addressline={Prime Video},
postcode={98109}, city={Seattle}, state={Washington}, country={USA}}
\affiliation[2]{organization={University of South Florida}, addressline={Department of Electrical Engineering},
postcode={33620}, city={Tampa}, state={Florida}, country={USA}}

\usepackage{prletters}
\usepackage{booktabs}
\usepackage{times}
\usepackage{epsfig}
\usepackage{graphicx}
\usepackage{amsmath}
\usepackage{amssymb}
\usepackage{xurl}
\usepackage{hyperref}

\begin{document}

\begin{frontmatter}

    \title{
  \mbox{}    
}

\begin{abstract}
While video action recognition has been an active area of research for several years, zero-shot action recognition has only recently started gaining traction. In this work, we propose a novel end-to-end trained transformer model which is capable of capturing long range spatiotemporal dependencies efficiently, contrary to existing approaches which use 3D-CNNs. Moreover, to address a common ambiguity in the existing works about classes that can be considered as previously unseen, we propose a new experimentation setup that satisfies the zero-shot learning premise for action recognition by avoiding overlap between the training and testing classes. The proposed approach significantly outperforms the state of the arts in zero-shot action recognition in terms of the the top-1 accuracy on UCF-101, HMDB-51 and ActivityNet datasets. The code and proposed experimentation setup are available in GitHub: https://github.com/Secure-and-Intelligent-Systems-Lab/SemanticVideoTransformer
\end{abstract}
    
  \end{frontmatter}
\vspace{10pt}
~


\input{Introduction}
\input{Related}

\input{Proposed}
\input{Experiments}
\input{Conclusion}

\bibliographystyle{elsarticle-num}
\bibliography{egbib}

\end{document}

%% file: Introduction.tex
\section{Introduction}
\label{sec:intro}

Several visual recognition tasks, such as image classification and video action recognition, have made tremendous progress in recent years, thanks to the availability of extensively annotated datasets and enhanced deep learning architectures. However, collecting and annotating video samples for every possible interaction between objects is impractical, therefore recognizing previously unseen actions remains a challenging task. On the other hand, humans are exceptionally good at recognizing new categories without seeing any visual samples. For example, if a person is familiar with \emph{ice skating} and understands the concept of \emph{dancing}, (s)he will have no trouble recognizing the action of \emph{ice dancing}. In the recent literature, this problem has drawn considerable attention and is known as Zero-Shot Learning (ZSL) for video action recognition. While several approaches have shown promising results in the image domain, zero-shot video action recognition remains largely unexplored.

\begin{figure}[t]
\vspace{-2mm}
\centering
\includegraphics[width=0.5\textwidth]{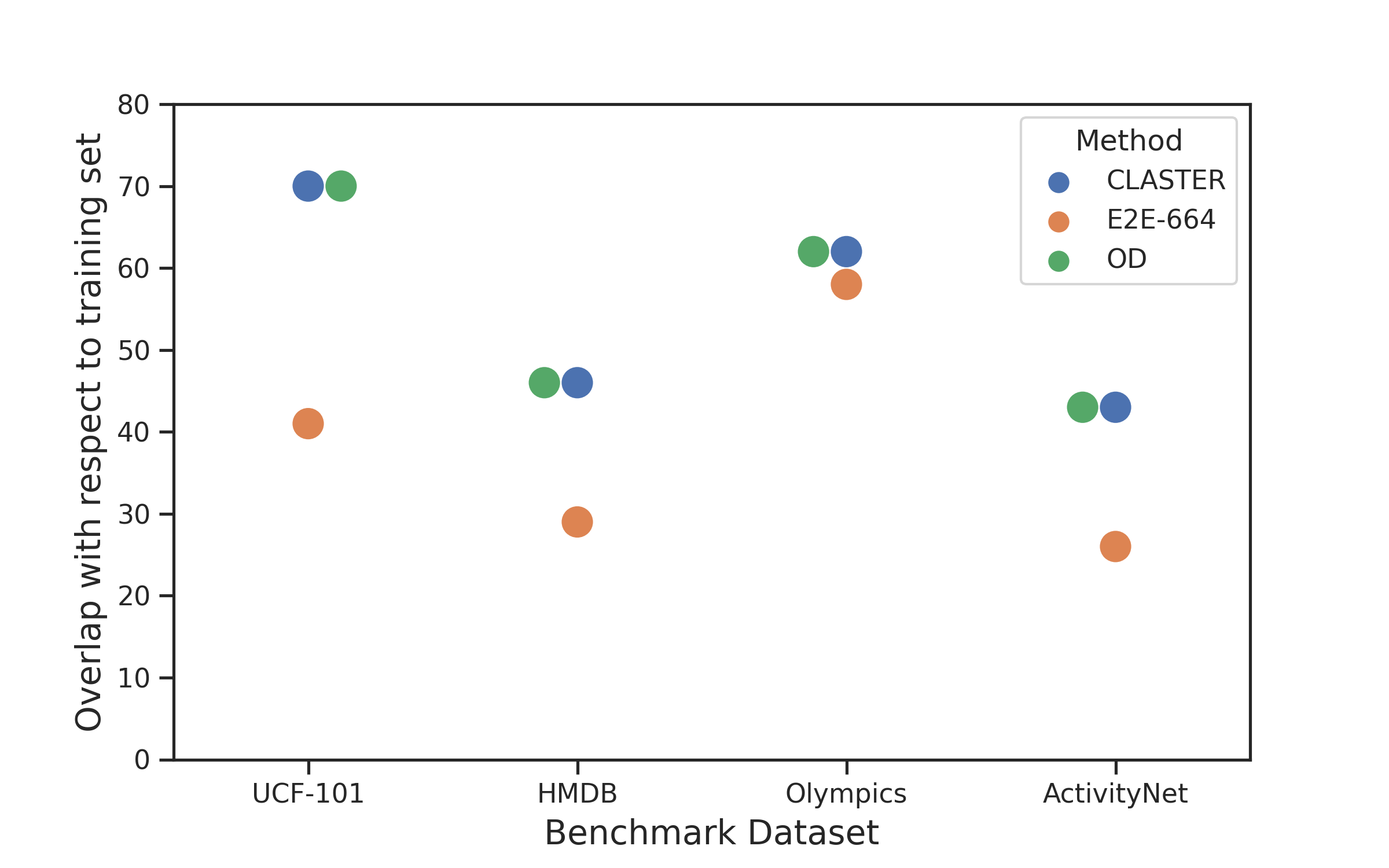}
\vspace{-5mm}
\caption{In the recent state-of-the-art approaches CLASTER \cite{gowda2021claster}, E2E-664 \cite{brattoli2020rethinking}, OD \cite{mandal2019out}, there is a significant overlap between the training classes and the testing classes. The vertical axis shows the percentage of test classes in the benchmark datasets (UCF-101, HMDB, Olympics, ActivityNet) that overlap with the training classes in the Kinetics dataset, which is significantly larger than the benchmark datasets used for testing. Since the test classes are supposed to be previously unseen, such high-percentage overlap (30\%-70\%) violates the ZSL premise.}
\label{f:ex_comp2}
\vspace{-3mm}
\end{figure}

In the existing research, ZSL is characterized as a classification problem in which a model is trained on a collection of known classes and then uses semantic attributes to identify unknown classes. Most existing approaches employ a 3D Convolutional Neural Network (3D-CNN) to extract visual features from videos. Because the utilized 3D-CNNs are pretrained on a variety of large-scale datasets, there is no obvious demarcation between what defines seen and unseen classes. Several recent techniques, for example, are pretrained on the Kinetics-400/600/700 \cite{kay2017kinetics,carreira2018short,smaira2020short} datasets and evaluated on the UCF-101 \cite{soomro2012ucf101}, HMDB-51 \cite{kuehne2011hmdb}, Olympics \cite{tang2012learning} and ActivityNet \cite{caba2015activitynet} datasets. However, as shown in Fig. \ref{f:ex_comp2}, several classes that are considered as ``unseen" are already present in the Kinetics dataset, which clearly violates the zero-shot paradigm. On the other hand, even a human is incapable of recognizing \emph{archery} as an activity if he has never seen a \emph{bow} and \emph{arrow} before and is oblivious to the concept of \emph{shooting}. Thus, it can be argued that it is redundant to include activities in the test set that are significantly dissimilar to the activities present in the training set. Furthermore, the existing evaluation setup consists of randomly splitting the test set classes into half and evaluating the proposed approach on that set. However, this causes an unfair comparison since modifying the split can significantly alter the result. To this end, we propose a new ZSL experimentation framework for action recognition, that addresses these issues and thus would provide guidance for future algorithm design. 

Since the majority of existing works employ 2D or 3D convolutions as the principal operator for learning spatiotemporal features from videos, they suffer from several challenges \cite{bertasius2021space}. First, inductive biases such as local connectivity, translation invariance, and a locally restricted receptive field severely limit the learning capability of convolutional models on large datasets. Moreover, convolutional kernels are incapable of capturing spatiotemporal relations that span multiple time instances \cite{bertasius2021space}. Finally, even with advances in hardware acceleration, training and evaluating deep CNNs on large video datasets remain computationally expensive. 

Motivated by these observations, we propose leveraging self-attention architectures, in particular a spatiotemporal transformer model, for extracting semantic embeddings from videos. The self-attention approach, in contrast to convolutional kernels, can capture long-range dependencies and is permutation invariant while being computationally efficient during training and inference. Our end-to-end trained transformer based approach is able to learn semantically more separable features as compared to 3D-CNNs.  
Our contributions can be summarized as follows:
\begin{itemize}
    \item \emph{A novel end-to-end transformer to learn visual-semantic representations.} To the best of our knowledge, this is the first work to propose a \emph{spatiotemporal transformer} for \emph{zero-shot video action recognition} although there are existing transformer methods for zero-shot image tasks.
    \item \emph{A thorough analysis of the shortcomings in the existing ZSL framework for action recognition.} We propose a new evaluation framework that satisfies the ZSL criterion for action recognition.
    \item \emph{An extensive evaluation of the proposed transformer method on several benchmark datasets.} The proposed method outperforms several state-of-the-art approaches by a wide margin under different experimental setups.  
\end{itemize}

%% file: Related.tex
\section{Related Works}
\label{s:related}

Video action recognition has been extensively studied over the past several years \cite{feichtenhofer2016convolutional,feichtenhofer2017spatiotemporal,wang2017spatiotemporal,fernando2015modeling,carreira2017quo,simonyan2014two}. In contrast, ZSL for video action recognition has only recently started gaining attention. Broadly, ZSL can be classified into the inductive setting \cite{action2vec,tarn,uar,zhang2018cross,piergiovanni2018learning}, where the test data is completely unknown during training, and the transductive setting \cite{xu2016multi,wang2017alternative,alexiou2016exploring,xu2015semantic,mishra2018generative,wang2017zero,xu2017transductive}, where the test data without class labels is available. In this work, we only focus on the inductive setting.  

Existing approaches have been predominantly dependent on word embeddings to tackle the problem of ZSL. Specifically, these approaches use a pretrained model to extract visual features from training videos and map them to a semantic space and hypothesize that a good robust generalization on the semantic space can lead to improved performance on unseen classes \cite{tarn,brattoli2020rethinking,gowda2021claster,action2vec,gan2015exploring,gan2016concepts,gan2016learning,zhang2018cross,xu2015semantic,xu2017transductive,xu2016multi}. For extracting the visual features, most recent approaches propose using a 3D-CNN, which takes 16 frames sampled from a video as input. In \cite{brattoli2020rethinking}, Brattoli et al. propose training a C3D \cite{tran2015learning} and a R(2+1)D model \cite{tran2018closer} in an end-to-end fashion for ZSL. On the other hand, Gowda et al. \cite{gowda2021claster} propose a reinforcement learning based clustering approach, which uses a two-stream I3D \cite{carreira2017quo} model for learning visual features. However, we argue that for realistic applications, sampling 16 frames from a video might not always be sufficient, especially when it comes to large-scale datasets. In \cite{carreira2017quo,he2016deep}, it is shown that increasing the number of input frames only marginally increases the performance of convolutional models. On the contrary, recent transformer based approaches \cite{bertasius2021space,zhang2021vidtr} have shown considerable performance gains when number of input frames are increased from 8 to 96, especially on tasks that require longer temporal reasoning. While earlier approaches use hand-crafted semantic features \cite{idrees2017thumos}, recent works have primarily use Word2Vec \cite{mikolov2013efficient} for generating the semantic embeddings from the class labels. However, such approaches are prone to the domain shift problem, which occurs when a model trained on the seen semantic labels is unable to generalize well to the unseen class labels \cite{action2vec}. 

Recently, Brattoli et al. \cite{brattoli2020rethinking} propose an extension on the work of Roitberg et al. \cite{roitberg2018towards} and formulate a novel evaluation protocol for satisfying the ZSL paradigm, which involves removing certain classes from the training set which overlap with the test set by using semantic embedding matching. However, in this paper, we show that such an approach fails to remove a significant portion of the overlapping classes, thus still violates the ZSL premise. Alternatively, Gowda et al. \cite{gowda2021new} propose a deterministic ``TruZe" split for the UCF-101 \cite{soomro2012ucf101} and HMDB-51 \cite{kuehne2011hmdb} datasets, by manually removing all classes which overlap with the Kinetics-400 dataset. While it is a promising approach, we show that it quickly becomes obsolete since several recent approaches use Kinetics-600/700 for training, which includes a majority of the classes from the ``TruZe" test set. Furthermore, neither of the proposed approaches remove classes which are significantly different from any of the classes in the training set. Hence, we propose a novel test set (Section \ref{s:fair}) which addresses all of the above mentioned issues.

%% file: Proposed.tex
\section{Proposed Approach}

\begin{figure}[tbh]
\centering
\includegraphics[width=0.45\textwidth]{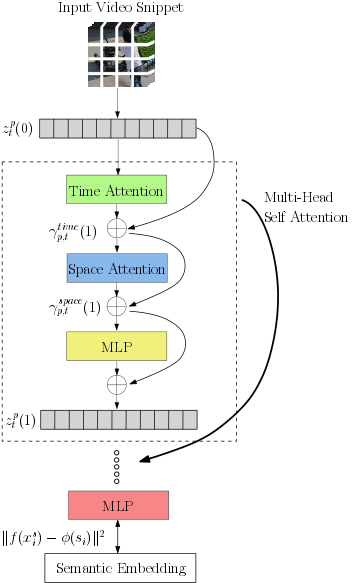}
\vspace{-2mm}
\caption{The proposed spatiotemporal transformer architecture, SVT, for video semantic embedding.}
\label{f:proposed_transformer}
\vspace{-3mm}
\end{figure}

In this section, we present the proposed approach for zero-shot video action recognition. We begin by carefully defining the problem formulation. Next, we introduce a novel end-to-end trained semantic video transformer (SVT), 
which leverages divided space-time attention to learn improved semantically differentiable visual features (Fig. \ref{f:proposed_transformer}). Then, we present a new data split for more suitable ZSL experimentation. Finally, we propose to enhance the available semantic embeddings using class descriptions instead of labels. 

\subsection{Problem Definition}

Traditionally, zero-shot action recognition has been defined as a classification problem, where given a training set of videos $X^s$ and labels $S$ from seen classes $\{(x^s_1,s_1), \dots, (x^s_N,s_N)\}$, we aim to accurately classify a set of videos  $X^u = \{x^u_1, \dots, x^u_M\}$ from previously unseen classes $U = \{u_1,\dots,u_M\}$, where $N$ and $M$ are the number of training and testing videos respectively. To satisfy the ZSL premise, there should be no overlap between the seen and unseen classes $(S \cap U = \varnothing)$. 

A broad generalization capability is needed to succeed in the ZSL action recognition problem. The challenge compared to the regular action recognition task lies in the fact that no direct mapping from the input videos to the output unseen class labels can be learned during training. Typically, semantic embeddings are used to bridge the input videos and the output unseen class labels, which consist of words. The idea behind this mainstream ZSL approach is to learn a semantic embedding model $f(x)$ for the input videos and choose the class that is semantically most similar. 


\begin{figure}[tbh]
\centering
\includegraphics[width=0.45\textwidth]{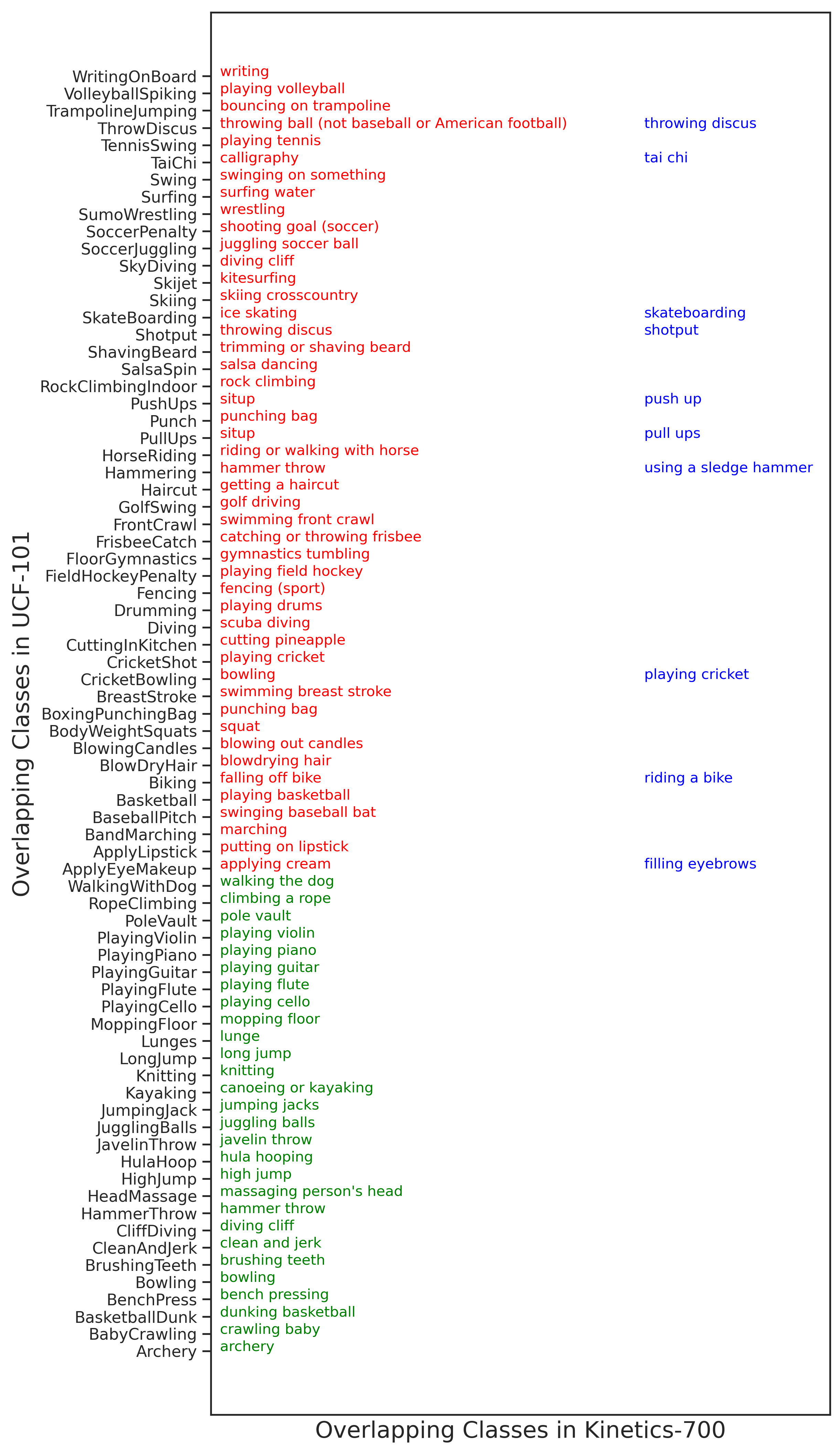}
\vspace{-2mm}
\caption{Visualization of the overlapping classes between UCF-101 and Kinetics-600/700. The Kinetics classes in green are considered as overlapping by the metric given by Eq. \eqref{eq:remove} and removed in \cite{brattoli2020rethinking}. The Kinetics classes in red are given by Word2Vec as the semantically nearest classes to the corresponding UCF-101 class, but not removed in \cite{brattoli2020rethinking} according to the criterion in Eq. \eqref{eq:remove}. For several cases, the actual closest Kinetics classes, shown in blue, are missed by Word2Vec. They are almost identical to the corresponding UCF-101 classes, and thus violate the ZSL idea.}
\label{f:overlap}
\vspace{-5mm}
\end{figure}

\subsection{Semantic Video Transformer (SVT)}

We consider the inductive ZSL approach for training our model, i.e., during training the model only has access to the videos and corresponding class labels from the seen classes. Most existing approaches extract visual features using pretrained 3D-CNNs. Recently, it was shown in \cite{brattoli2020rethinking} that end-to-end training a 3D-CNN performs significantly better than directly using a pretrained 3D-CNN model. However, the inability of 3D-CNNs to capture spatiotemporal information over a long time span make them unsuitable for large scale datasets such as Kinetics and ActivityNet. Hence, we propose a novel end-to-end trained transformer model, called 
\emph{Semantic Video Transformer (SVT)}. 
The overall structure of SVT is shown in Fig. \ref{f:proposed_transformer}, which is based on the recently proposed TimeSformer \cite{bertasius2021space} architecture. Specifically, we leverage the self-attention blocks from the Vision transformer (ViT) \cite{dosovitskiy2020image} model and space-time attention from the TimeSformer model. However, contrary to the existing video transformers, we train our model to learn visual-semantic representations.


\textbf{Input:} Even with a computationally efficient architecture, end-to-end training on the entire video is impractical due to GPU limitations. Following the existing video classification approaches, we first sample a clip $y$ of $F$ frames and size $H \times W$ from the input video $x$. The standard NLP transformer architecture \cite{vaswani2017attention} requires a 1D sequence of token embeddings as an input. Hence, as shown in Fig. \ref{f:proposed_transformer}, we first breakdown the entire video clip $y$ into a sequence of 2D patches, $e_t^p \in \mathbb{R}^{3 \times P^2}$, where $p=1,\dots,N$ represents the spatial locations (i.e., patch index), $t=1,\dots,F$ denotes the temporal index, $3$ is the number of color channels, and $P \times P$ is the patch size. 
Next, we flatten each patch $e^p_t$ into $v_t^p \in \mathbb{R}^{3P^2}$ and linearly map it into a score vector using a trainable linear projection $E \in \mathbb{R}^{q \times 3P^2}$:
\begin{equation}
\label{eq:linear}
    z^p_t(0) = Ev_t^p + \mu^p_t,
\end{equation}
where $\mu^p_t \in \mathbb{R}^{q}$ is a latent vector learned to encode the spatiotemporal position of each $(p,t)$ pair. Following the NLP transformer BERT \cite{devlin2018bert} we add a latent vector $z^0_0(0) \in \mathbb{R}^q$ for an additional fictional patch which will be learned to represent the score vector for the entire video by interacting with all patches through time and space self-attention. $\{z^0_0(0), z^p_t(0)\}_{p,t}$ is the input to the transformer model.  

\textbf{Overall Structure:} The transformer model consists of $L$ sequential encoding blocks, each of which includes $A$ parallel self-attention heads. The first encoding block processes the input $\{z^0_0(0), z^p_t(0)\}$ in parallel in its self-attention heads, as explained next, and outputs $\{z^0_0(1), z^p_t(1)\}$ to the second encoding block for each patch $(p,t)$. Similarly, each block $l$ gets $\{z^0_0(l-1), z^p_t(l-1)\}$ and outputs $\{z^0_0(l), z^p_t(l)\}$. Finally, the output of the last block $z^0_0(L)$ is used to obtain the semantic embedding for the video. The role of $z^0_0(L)$ is to learn an effective summary score for the entire video, as an alternative to the straightforward approach of simply averaging $z^p_t(L)$ over all patches and frames. 
$z^0_0(L)$ is passed through an MLP with three hidden layers and ReLU activation function to obtain the semantic embedding $f(x)$ for video $x$. 

\textbf{Training and Inference:} The whole transformer model is trained end-to-end by minimizing the loss function 
\begin{equation}
    C = \|f(x^s_i)-\phi(s_i)\|^2,
\end{equation}
where $\phi(s_i)$ is the semantic embedding of the class label/description $s_i$ for the training video $x_i^s$ from a Sent2Vec model \cite{pgj2017unsup}. 
After the model is trained, for ZSL inference, the semantic embedding of the test video $f(x^u_i)$ is found using the transformer model, and the class closest to the video in the semantic space is chosen, $$\arg\min_j D_{\cos}(f(x^u_i),\phi(u_j)),$$ where $D_{\cos}$ is the cosine distance and $\phi(u_j)$ is the Sent2Vec semantic embedding of the class label/description $u_j$. 

\textbf{Class Descriptions:} 
Due to the unavailability of class descriptions or attributes, existing methods in the literature alternatively use the class labels and Word2Vec to extract the semantic embeddings $\phi(u_j)$. However, we argue that such an approach would severely limit the performance of a model. Specifically, several class labels in all the datasets are not sufficiently distinctive and could refer to various different activities. For example, \emph{diving} in UCF-101 could either refer to \emph{scuba diving}, \emph{cliff diving}, \emph{sky diving} or \emph{springboard diving}. Furthermore, using Word2Vec on multi-word labels is not an efficient technique since averaging the semantic embedding over several words might lose the context. This can be also seen in Fig. \ref{f:overlap}, where the semantic embedding of several multi-word class labels is mapped incorrectly. To circumvent these issues, we first manually annotate all the training and testing datasets with one line class descriptions. We then use a Sent2Vec \cite{pgj2017unsup} model to better capture the semantic information with respect to the context of the class description sentence. The annotated class descriptions are available in GitHub \footnote{\url{https://github.com/Secure-and-Intelligent-Systems-Lab/SemanticVideoTransformer}}. In Section \ref{s:ablation}, we evaluate the contribution of class descriptions by comparing the performance of the proposed SVT method with class labels and descriptions in an ablation study. 

\subsection{Fair Zero-Shot Action Recognition} 
\label{s:fair}
While several recent works have shown promising results on the benchmark datasets, there are still several shortcomings. In contrast to earlier works, recent approaches \cite{brattoli2020rethinking,gowda2021claster,gowda2021new,roitberg2018informed} use large-scale external datasets such as the Kinetics 400/600/700 dataset to pretrain the visual feature extractors. However, as shown in Fig. \ref{f:ex_comp2}, there is a significant overlap between the seen and unseen classes, $S\cap U \neq \varnothing$, which clearly violates the ZSL paradigm. Recently, Brattoli et al. \cite{brattoli2020rethinking} proposed a novel training protocol which involves removing classes from the training set if
\begin{equation}
\label{eq:remove}
   \min_{s_i \in S, u_j \in U}  D_{\cos}(\phi(s_i),\phi(u_j)) < \tau,
\end{equation}
where $\tau$ is set as $0.05$. However, in Fig. \ref{f:overlap}, we show that even after applying such a constraint, there are several overlapping classes which are not removed. This can be partly attributed to semantic mismatch due to domain shift, where two very similar classes are called by slightly different names. For example, \emph{blowing out candles} and \emph{blowing candles} refer to the same class and yet the cosine distance between them in the semantic space is much greater than $\tau$. Moreover, removing overlapping classes from the training set is a cumbersome approach since it requires retraining computationally expensive models for every individual test set. On the contrary, it would be far easier to remove overlapping classes from the test set as the class information is already available in the training data. 

Furthermore, it is also worth considering whether a given test class is completely irrelevant with respect to all the seen classes. Intuitively, even humans cannot comprehend a new activity if it involves interactions and objects that are never seen before. Hence, we also propose removing classes which are significantly different from the activities present in the training dataset. While keeping such classes does not necessarily violate the ZSL premise, it introduces a source of randomness since no reasonable algorithm can be expected to recognize such activities. 

In the current literature, most approaches randomly split a single dataset and evaluate performance on it over several trials. Such an evaluation setup is not practical in a real-world scenario since most of the videos would have disjoint sources. Hence, to ensure a realistic ZSL setting, we extend the setup discussed in \cite{brattoli2020rethinking}, which suggests using independent datasets for training and testing. However, \cite{brattoli2020rethinking} does not fully account for overlapping classes between the train and test sets, as seen in Fig. \ref{f:overlap}. We propose a new ZSL experimentation setup for action recognition, called \emph{Fair ZSL}. For each dataset, completely removing classes that either overlap or completely irrelevant with respect to the train set (Kinetics 600/700) leaves us with a very small number of test classes per test set (as shown for UCF-101 in Fig. \ref{f:overlap}). Hence, we propose pooling the valid test classes from all benchmark datasets to form a novel test set. In the proposed Fair ZSL setup, there are 30 unique classes\footnote{The detailed list of all the classes in the proposed split is available at \url{https://github.com/Secure-and-Intelligent-Systems-Lab/SemanticVideoTransformer}.} from the UCF-101, HMDB-51, and ActivityNet datasets, as shown in Table \ref{tab:fzsl}. Each class was carefully handpicked such that it does not violate the ZSL premise. Additionally, the proposed test set is also more robust since it evaluates how well a single model accounts for domain shifts in addition to the ZSL performance. 

\begin{table}[t]
    \centering
    \begin{tabular}{lc}  
    \toprule
     Benchmark Dataset & Selected Classes \\
    \toprule
    UCF-101 & 8 \\
    HMDB-51 & 3  \\
    ActivityNet-101  & 19 \\
    \bottomrule
    \end{tabular}
    \caption{Number of classes from each benchmark dataset in the proposed test set for Fair ZSL.}
    \label{tab:fzsl}
    \vspace{-3mm}
\end{table}

\vspace{-1mm}
\subsection{Implementation Details}

The proposed SVT model is built upon the PySlowFast \cite{chen2019drop} package. For training the model, the shorter side of the input video is first resized to 256 pixels and then randomly cropped to form a $224 \times 224$ ($H \times W$) video snippet. The patch size is chosen as $16 \times 16$, resulting in $N=196$ patches in a frame. The size of the score vectors ($z^p_t(l)$) at each encoding block for each patch is $q=768$, and the number of self-attention heads is $A=12$. The size of the learned semantic embedding $f(x)$ and the Sent2Vec embedding for class descriptions is $600$. We train two versions of the proposed transformer, SVT-8 and SVT-96 for $F=8$ and $F=96$ frames in the input video snippet, respectively. In all our experiments, we use the SVT-96 model unless explicitly stated otherwise. We train the models on 4 NVIDIA A40 GPUs with a batch size of 16 for SVT-8 and 4 for SVT-96. The loss function is minimized via synchronized SGD with a learning rate of 0.002. To extract semantic embeddings, we use the Sent2Vec algorithm proposed in \cite{pgj2017unsup}.      

\subsection{Computational Efficiency} 

Thanks to the scalability of the proposed SVT model, we are able to vary the length of the input video snippet (i.e., number of frames $F$), which also leads to an increase in the number of input tokens. In Table 3, we see a significant increase in the performance when the number of input frames are increased from 8 to 96. Increasing the number of video frames is intuitive since it allows a model to better capture the spatiotemporal activities that span several frames. However, due to the current GPU limitations, we are unable to further increase the input length. On the other hand, even after increasing our model complexity to accommodate 96 input frames, our model is still more computationally efficient as compared to the I3D model with 8 input frames, which requires 10.8 TFLOPS for inference, in contrast to the proposed SVT-8 model, which only requires 0.79 TFLOPS, and SVT-96, which requires 7.57 TFLOPS.  

%% file: Experiments.tex
\vspace{-1mm}
\section{Experiments}


\subsection{Datasets}

Most of the recent works evaluate their performance on three publicly available benchmark datasets, namely the UCF-101, HMDB-51 and Olympics dataset. The UCF-101 dataset consists of 13,320 videos from 101 classes, primarily focusing on five types of actions. The HMDB-51 dataset consists of 6767 videos from 51 classes based on daily human actions. The Olympics dataset consists of 16 categories, related to an Olympic sport. Recently, Brattoli et al. \cite{brattoli2020rethinking} evaluated the performance of their approach on the ActivityNet dataset by extracting labelled frames from every video. As compared to the other benchmark datasets, ActivityNet is considerably more comprehensive, consisting of 27,801 videos from 200 classes related to daily activities. The Kinetics-700 dataset is the largest dataset available for training video action recognition models, with more than 500K videos in 700 categories sourced from YouTube. Since several classes from Kinetics-700 were not available or had files corrupted, we use the Kinetics-600 dataset in our experimental setup. Due to its small size and significant overlap with Kinetics, we do not consider the Olympics dataset in our evaluations.      

\subsection{Experimental Setup}
\label{s:exp_setup}
To analyze the performance of our proposed SVT model and provide a fair comparison with benchmark algorithms, we first follow the training and evaluation protocols used in the existing papers. The existing training protocols can be broadly classified into two categories, open-ended and restrictive. Most existing works are based on the open-ended formulation, whereas a few recent approaches \cite{brattoli2020rethinking,roitberg2018towards,gowda2021new} use the restrictive one.

\textbf{Open-Ended:} In this setup, a model is first trained on a large-scale visual dataset such as the Kinetics dataset without removing any classes, and then evaluated on a smaller application specific dataset such as UCF or HMDB. A few approaches also further fine-tune their models on the smaller dataset by randomly splitting it into train and test sets; however doing so does not necessarily improve  performance. Moreover, as also suggested in \cite{brattoli2020rethinking}, a model ideally should have separate video sources for training and testing to evaluate its generalization capability.     

\textbf{Restrictive:} In the restrictive training approach proposed in \cite{brattoli2020rethinking}, we remove all classes from Kinetics-600 whose distance to any class in UCF $\cup$ HMDB is smaller than $\tau$ when testing on UCF or HMDB, which results in a subset of Kinetics with 564 classes. For testing on ActivityNet, an even more restrictive approach is proposed which involves removing all classes whose distance to any class in ActivityNet $\cup$ UCF $\cup$ HMDB is smaller than $\tau$. This setting leads to an even smaller subset of Kinetics with 505 classes.    

For the open-ended formulation, we train our model on the entire Kinetics-600 dataset and choose to forgo fine-tuning on UCF or HMDB since transformer based models require significant amount of data to learn meaningful representations. To make our evaluation comparable to the existing approaches, we first randomly split the test dataset in half and evaluate our proposed approach on it over 10 trials. For the restrictive approach, in addition to the random split, we also evaluate our model on the entire UCF and HMDB datasets (Table \ref{tab:sota2}) since it allows for a more robust evaluation due to lack of randomness.   

\textbf{Fair ZSL:} Since neither of the existing training protocols provide the true essence of ZSL, we propose a new setup where a model is trained on the entire Kinetics-600/700 dataset and evaluated on the proposed test set discussed in Section \ref{s:fair}. For evaluation, we compare our approach in Table \ref{tab:sota3} with the recently proposed E2E model \cite{brattoli2020rethinking} and our implementation of the CLASTER model \cite{gowda2021claster}. Recently, \cite{chen2021elaborative} proposed an elaborate rehearsal approach which, in addition to a 3D-CNN, explicitly uses an object classifier trained on ImageNet to learn objects detected in videos. Since their approach and problem setup is significantly different from the existing approaches, we do not compare our approach with them.

\subsection{Results}

In the model name ``SVT-X(Y) + Z", X refers to the number of frames used in the analysis of each video, Y denotes the number of the classes from Kinetics-600 used in training, and Z represents the data type used for class semantic embedding (CD for class descriptions and CL for class labels). The impacts of frame number and class descriptions are analyzed in Section \ref{s:ablation}.

We first compare the proposed method with the state-of-the art approaches under the dominant random split setup in Table \ref{tab:sota}. However, \cite{ghosh2021learning} have presented their results using a different split (e.g., 78-23 for UCF-101), so we do not include their results in Table \ref{tab:sota}. The test datasets are randomly split into half using the seed 10, as in \cite{brattoli2020rethinking}. It is seen that the proposed spatiotemporal transformer based ZSL approach consistently outperforms all other state-of-the-art approaches under all settings. On the UCF-101 dataset, we notice a significant improvement of 21.9\% and 5.2\% over the next best results, for the OE and R protocols, respectively. On the HMDB-51 dataset, the improvements of the proposed method are 3.4\% and 1.8\% for the OE and R protocols, respectively. None of the existing approaches present their results on the ActivityNet dataset for the OE protocol, hence we could not compare our result on it. However, for the R protocol, we outperform the E2E approach by a wide margin of 9.2\%. 

\begin{table}[t]
    \setlength{\tabcolsep}{2pt}
    \centering
    \resizebox{0.48\textwidth}{!}{%
    \begin{tabular}{lcccc}  
    \toprule
    Method & Protocol & UCF & HMDB & ActivityNet \\
    \toprule
    DataAug~\cite{xu2016multi} & OE & 18.3 & 19.7 & -  \\
    InfDem~\cite{roitberg2018informed} & OE & 17.8 & 21.3 & - \\
    Bidirectional~\cite{wang2017zero} & OE & 21.4 & 18.9 & - \\
    TARN~\cite{tarn} & OE & 19 & 19.5 & - \\
    Action2Vec~\cite{action2vec} & OE & 22.1 & 23.5 & - \\
    OD~\cite{mandal2019out} & OE & 26.9 & 30.2 & - \\
    CLASTER~\cite{gowda2021claster} & OE & 46.4 & 36.8 & - \\
    DASZL~\cite{kim2021daszl} & OE & 48.9 & - & - \\
    GGM~\cite{mandal2019out} & OE & 20.3 & 20.7 & - \\
    \textbf{SVT-96(600) + CD (Ours)} & OE & \textbf{68.3} & \textbf{40.2} & \textbf{44.8 }\\
    \midrule
    E2E (605classes) & R & 44.1 & 29.8 & 26.6 \\
    E2E (664classes) & R & 48 & 32.7 & - \\
    PS-ZSAR (662 classes) \cite{kerrigan2021reformulating} & R & {49.2}  & {33.8} & - \\
    \textbf{SVT-96(505) + CD (Ours)} & R & {45.6} & {31.3} & \textbf{35.8} \\
    \textbf{SVT-96(564) + CD (Ours)} & R & \textbf{53.2} & \textbf{34.5 }& - \\
    \bottomrule
    \end{tabular}
    }
    \caption{Comparison with the state-of-the-art methods on standard benchmark datasets using the open-ended (OE) and Restrictive (R) protocols. All the methods are evaluated by randomly splitting the dataset in half and averaging the results over 10 trials.}
    \label{tab:sota}
    \vspace{-1mm}
\end{table}

A disjoint training and testing split allows us to evaluate our model on the entire UCF-101, HMDB-51 and ActivityNet datasets without any random split. As shown in Table \ref{tab:sota2}, the performances drop compared to the random split case in Table \ref{tab:sota}. However, we still considerably outperform the E2E based approach under the R protocol while using the class descriptions and class labels. Since the rest of the existing state-of-the-art approaches use some part of UCF-101 or HMDB-51 for fine tuning their models, we could not compare our results with them in this setup. 

Finally, we also consider the Fair ZSL protocol proposed in Section \ref{s:exp_setup}. In Table \ref{tab:sota3}, we compare our model with the E2E \cite{brattoli2020rethinking} method and our implementation of the CLASTER \cite{gowda2021claster} method. We again notice a significant improvement of 5.9\% over E2E, which shows the robustness and generalization capability of the proposed method. It should be noted that we use the E2E model trained on 664 classes for comparison since the model trained on all 700 classes is not available. However, since we train on Kinetics-600, which has even less number of classes, we believe it is a fair comparison.  

\begin{table}[t]
     \setlength{\tabcolsep}{2pt}
     \centering
    \resizebox{0.5\textwidth}{!}{%
    \begin{tabular}{lccccc}  
    \toprule
     Method & Embedding Type & Protocol & UCF & HMDB & ActivityNet\\
    \toprule
    E2E (605classes) & CL & R & 35.3 & 24.8 & 20.0\\
    E2E (664classes) & CL & R & 37.6 & 26.9 & - \\
    \textbf{SVT-96(505)} & CD & R & {36.1} & {25.7} & \textbf{26.3} \\
    \textbf{SVT-96(564)} & CD & R & \textbf{40.2} & \textbf{30.5 } & - \\
    \textbf{SVT-8(564)} & CD & R & 37.6 & 27.9 & - \\
    \textbf{SVT-96(505)} & CL & R & {33.9} & {25.2} & 24.7\\
    \textbf{SVT-96(564)} & CL & R & {38.3} & {27.6 } & - \\
    \bottomrule
    \end{tabular}
    }
    \caption{Comparison with the E2E \cite{brattoli2020rethinking} approach under the Restrictive (R) protocol using the entire datasets for testing without any random split. E2E and SVT are trained on Kinetics-700 and Kinetics-600, respectively.}
    \label{tab:sota2}
\end{table}

\begin{table}[t]
    \setlength{\tabcolsep}{3pt}
    \centering
    \begin{tabular}{lccc}  
    \toprule
     Method & Protocol & FZSL Split \\
    \toprule
    CLASTER & FZSL & 24.3 \\
    E2E (664 classes) & FZSL & 30.8  \\
    \textbf{SVT-96(600) + CD} & FZSL & \textbf{36.7} \\
    \bottomrule
    \end{tabular}
    \caption{Comparison with E2E \cite{brattoli2020rethinking} and our implementation of CLASTER \cite{gowda2021claster} using the proposed Fair ZSL protocol. The models are tested on the test set proposed in Section \ref{s:fair}  (Table \ref{tab:fzsl}).}
    \label{tab:sota3}
    \vspace{-1mm}
\end{table}

\subsection{Ablation Study}
\label{s:ablation}

In this section, we analyze the contributions of different components of the proposed approach by performing empirical studies. 

\begin{figure}[tbh]
\centering
\includegraphics[width=0.5\textwidth]{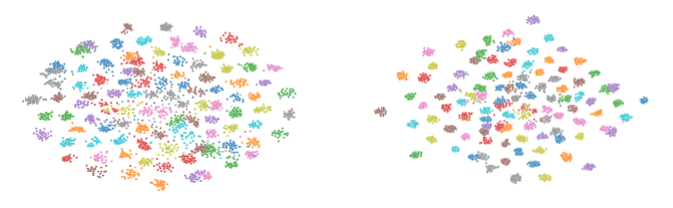}
\caption{t-SNE visualization of the I3D (left) and SVT (right) features extracted from the UCF-101 dataset. Each point represents a video and various classes are represented with different colors. We see that the features learned by the proposed SVT model (right) are semantically more separable than the I3D features (left). Best viewed in color.}
\label{f:tsne}
\vspace{-1mm}
\end{figure}

\textbf{Impact of the Spatiotemporal Transformer:} Here, we study how well the proposed SVT model is able to learn the spatiotemporal visual features. For comparison, we consider the a state-of-the-art 3D-CNN model called I3D \cite{carreira2017quo}, which has been a popular choice for video action recognition, and has been used by several existing approaches \cite{mandal2019out,gowda2021claster,xian2018feature} for zero-shot action recognition. For SVT, we extract the output of the last block $z_0^0(L)$, which serves as a visual embedding for the entire video. To extract I3D features, we follow the approach employed by \cite{mandal2019out,gowda2021claster,xian2018feature} and average the output from the \emph{Mixed 5c} layer across the temporal dimension followed by pooling by four in the spatial dimension and finally flattening it to a vector of size 4096. We use both RGB and flow features and concatenate them to form a vector of size 8192. In Fig. \ref{f:tsne}, we present the t-SNE visualization for the I3D and SVT features on UCF-101, where each point represents a video in the UCF dataset. It is clearly seen that, as compared to the I3D features, SVT learns more semantically separable features with more well-defined clusters. We also quantitatively compare them in Table \ref{tab:quant} by using various statistical metrics. The average silhouette score measures how tightly grouped all the points in the cluster are. The adjusted rand index computes a similarity measure between the clusterings and the ground truth. The homogeneity score checks if a cluster contains samples belonging to a single class. Finally, we also apply a simple $k$-NN classification algorithm on the extracted features to compute the accuracy for traditional video classification. In all metrics, the SVT features provide a better separation.

\begin{table}
    \setlength{\tabcolsep}{2pt}
    \centering
    \begin{tabular}{lccc}  
    \toprule
     Method & I3D & SVT &  \\
    \toprule
    Average Silhouette & 0.119 & \textbf{0.196} \\
    Adjusted Rand Index & 0.80 &\textbf{ 0.88}  \\
    Homogeneity Score & 0.92 & \textbf{0.96} \\
    Classification & 0.93 & \textbf{0.96} \\
    \bottomrule
    \end{tabular}
    \caption{Comparison between the SVT and I3D features in terms of clustering and classification performance using different metrics. The classification accuracy is for the $k$-NN classifier.}
    \label{tab:quant}
\end{table}

\textbf{Impact of Class Descriptions:} In Table \ref{tab:cl-cd}, we compare the performance of our approach when it uses class labels (CL) and class descriptions (CD) for semantic embedding. Specifically, we train the SVT model on the semantic embeddings extracted using the Sent2Vec model on class labels/descriptions, and test on the entire UCF, HMDB and ActivityNet datasets using the restrictive (R) setting. We see that there is a noticeable improvement in the performance when class descriptions are used, which ascertains our conjecture that descriptive semantic embeddings are crucial in improving the learned class representations. Moreover, our performance using class label embeddings is still better than existing state-of-the-art, which shows the efficacy of our proposed SVT model. While generating class descriptions require some human involvement, we argue that the cost is still significantly less as compared to video level annotations required for supervised learning. Furthermore, manually defined attributes are already being used for zero-shot image classification \cite{huynh2020fine} since class labels are not always discriminative enough to distinguish the context.

\begin{table}[t]
    \setlength{\tabcolsep}{2pt}
    \centering
   
   \begin{tabular}{lcccc}  
   \toprule
    Method & Embedding Type & UCF & HMDB & ActivityNet \\
   \toprule
   {SVT-96(505)} & CL & {33.9} & {25.2} & 24.7\\
   {SVT-96(564)} & CL & {38.3} & {27.6 } & - \\   
   {SVT-96(505)} & CD & {36.1} & {25.7} & 26.3\\
   {SVT-96(564)} & CD & {40.2} & {30.5} & - \\
   \bottomrule
   \end{tabular}
   \caption{Comparison between the performance of SVT while using class labels (CL) and class descriptions (CD) under the restrictive protocol.}
   \label{tab:cl-cd}
\end{table}



\textbf{Number of Input Frames:} Finally, we analyze the impact of the number of input frames on the SVT performance. As shown in Table \ref{tab:sota2}, there is a 2.6\% gain in both UCF and HMDB datasets when 96 frames are used instead of 8. 

%% file: Conclusion.tex
\section{Conclusion}

In this work, we introduce a spatiotemporal transformer architecture for zero-shot video action recognition, called SVT. Moreover, we highlight several areas where the existing approaches either violate the zero-shot learning (ZSL) premise or are unable to perform well due to the limited capabilities of 3D-CNN based visual extractors. We propose a new evaluation protocol, Fair ZSL, that strictly adheres to the ZSL premise. Through several experiments, we show that the proposed approach consistently outperforms the existing approaches under various experimental setups, including the existing ones in the literature and the proposed Fair ZSL setup. 